\documentclass{article}
\usepackage{spconf,amsmath,graphicx}
\usepackage{cite}
\usepackage{booktabs}
\usepackage{bm}
\usepackage{amsfonts}
\usepackage{makecell}
\usepackage{colortbl}
\usepackage{soul}
\usepackage[english]{babel}
\usepackage{comment}
\usepackage{xurl}
\usepackage{stfloats}

\sethlcolor{E6E5FD}

\usepackage{enumitem}
\setlist{nosep, leftmargin=14pt}

\usepackage{mwe} 

\usepackage{xcolor}
\definecolor{skyblue}{HTML}{497BB8}

\usepackage[colorlinks=true, 
urlcolor=violet, 
linkcolor=blue, 
citecolor=skyblue]{hyperref}

\addto\extrasenglish{%
}

\addto\extrasenglish{%
}

\newcommand{\rr}{\color{red}} 
\newcommand{\bb}{\color{black}}
\newcommand{\cc}{\color{cyan}} 

\definecolor{lightblue}{HTML}{E6E5FD}

\title{
Context-Gated Cross-Modal Perception with Visual Mamba for PET-CT Lung Tumor Segmentation
}

\name{%
  \parbox{\textwidth}{\centering
  Elena Mulero Ayllón$^{\dagger}$, Linlin Shen$^{\P}$, Pierangelo Veltri$^{\phi}$, Fabrizia Gelardi$^{\ast \Vert}$\\
  Arturo Chiti$^{\ast \Vert}$, Paolo Soda$^{\dagger \ddagger}$, and 
  Matteo Tortora$^{\S \star}$\thanks{$^{\star}$ Corresponding author: \url{matteo.tortora@unige.it}}
  }
}

\address{%
  \parbox{0.92\linewidth}{\centering\footnotesize
  $^{\dagger}$ Unit of Artificial Intelligence and Computer Systems, Università Campus Bio-Medico di Roma, Italy\\
  $^{\P}$ College of Computer Science and Software Engineering, Shenzhen University, China\\
  $^{\phi}$ Dept. of Computer Engineering, Modeling, Electronic and System Engineering, University of Calabria, Italy\\
  $^{\ast}$ IRCCS San Raffaele Hospital, Italy\\
  $^{\Vert}$ Faculty of Medicine, Vita-Salute San Raffaele University, Italy\\
  $^{\ddagger}$ Dept. of Diagnostics and Intervention, Radiation Physics, Biomedical Engineering, Umeå University, Sweden\\
  $^{\S}$ Dept. of Naval, Electrical, Electronics and Telecommunications Engineering, University of Genoa, Italy
  }
}

\begin{document}
\maketitle
\begin{abstract}
Accurate lung tumor segmentation is vital for improving diagnosis and treatment planning, and effectively combining anatomical and functional information from PET and CT remains a major challenge. 
In this study, we propose vMambaX, a lightweight multimodal framework integrating PET and CT scan images through a Context-Gated Cross-Modal Perception Module (CGM). 
Built on the Visual Mamba architecture, vMambaX adaptively enhances inter-modality feature interaction, emphasizing informative regions while suppressing noise. 
Evaluated on the PCLT20K dataset, the model outperforms baseline models while maintaining lower computational complexity. 
These results highlight the effectiveness of adaptive cross-modal gating for multimodal tumor segmentation and demonstrate the potential of vMambaX as an efficient and scalable framework for advanced lung cancer analysis.
The code is available at~\url{https://github.com/arco-group/vMambaX}.
\end{abstract}
\section{Introduction}
\label{sec:intro}
Lung cancer is a leading cause of cancer-related deaths worldwide, where early detection and accurate assessment are essential to improving treatment planning and patient outcomes~\cite{tortora2021deep}. 
Medical imaging plays a crucial role in this process, offering detailed anatomical and functional insights into pulmonary lesions. 
Consequently, automated lung tumor segmentation has become fundamental for delineating tumor boundaries, enabling quantitative analysis, and assisting clinical decision-making.

Over the years, segmentation methods have evolved from traditional image processing to deep learning-based approaches~\cite{ayllon2025can}, with convolutional neural networks (CNNs) and transformer architectures achieving remarkable results. 
Different imaging modalities, such as computed tomography (CT) and positron emission tomography (PET), provide complementary information on tumor morphology and metabolism.

To exploit these complementary strengths, multimodal segmentation models have emerged as a promising direction~\cite{basu2024systematic}. 
In lung cancer, CT and PET are particularly relevant: CT offers high-resolution anatomical detail, while PET captures metabolic activity, enabling a more comprehensive tumor characterization when jointly analyzed.
Despite significant advances, existing multimodal segmentation approaches face key challenges. Many involve high computational costs, limiting their clinical applicability. Moreover, current fusion strategies often fail to fully exploit the complementary nature of different modalities, leading to suboptimal feature integration and reduced accuracy. These limitations highlight the need for an efficient, modality-aware framework that selectively integrates anatomical and functional information for improved tumor delineation.

In this study, we present \textbf{vMambaX}, a lightweight multimodal framework that leverages the Visual Mamba architecture to integrate complementary PET and CT information for accurate lung tumor segmentation.
Our main contributions are summarized as follows:
\begin{itemize}
\item We propose a multimodal PET-CT segmentation model that effectively combines anatomical and metabolic information to improve tumor segmentation in lung cancer.
\item We introduce a \emph{Context-Gated Cross-Modal Perception} mechanism that learns adaptive, modality-specific gating functions to emphasize anatomically and functionally relevant features while suppressing modality-specific noise.
\item We evaluate on the PCLT20K dataset, demonstrating that our approach achieves state-of-the-art segmentation performance with substantially lower computational complexity than existing baselines.
\end{itemize}

\section{Related Works}
\label{sec:related_work}
Multimodal learning integrates complementary information from heterogeneous sources to enhance robustness and predictive accuracy~\cite{tortora2023radiopathomics}, a crucial advantage in medical imaging for achieving precise and reliable segmentation. Among the various modality combinations in oncological imaging, PET-CT integration is particularly effective, as each modality provides complementary information.

Deep learning has transformed medical image segmentation, with convolutional neural networks (CNNs) achieving state-of-the-art performance by learning hierarchical representations from multimodal data. U-Net and its variants, in particular, have demonstrated high accuracy due to their encoder–decoder structure and skip connections that preserve spatial detail~\cite{azad2024medical}. More recently, transformer-based models, such as Vision Transformers, Swin-Transformers, and hybrid CNN-Transformer networks, have emerged as powerful alternatives capable of capturing long-range dependencies and global context often missed by CNNs~\cite{xiao2023transformers}.

Recent multimodal fusion strategies further aim to exploit cross-modality relationships through learnable, attention-guided mechanisms that dynamically weight modalities based on their contextual relevance. These adaptive approaches have consistently improved PET-CT segmentation accuracy, underscoring the importance of integrating complementary anatomical and functional information within unified frameworks.

\section{Methods}
\label{sec:methods}

\begin{figure*}
\includegraphics[width=\textwidth]{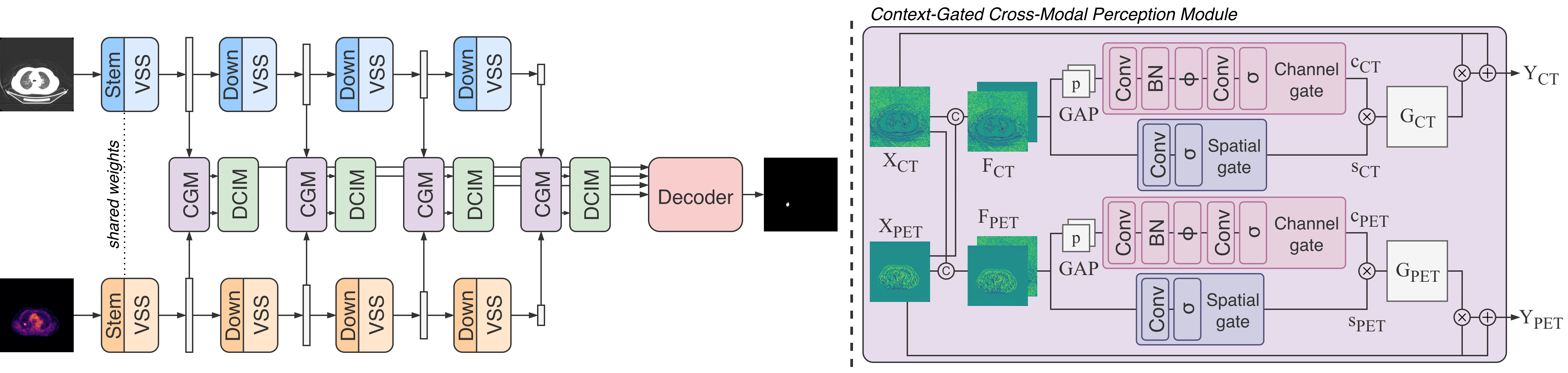}
\centering
\caption{
Overview of the proposed vMambaX architecture (left) and detailed structure of the CGM mechanism (right).}
\label{fig:overall_method}
\end{figure*}

As illustrated in~\autoref{fig:overall_method}, the proposed vMambaX framework adopts a dual-branch design composed entirely of state-space modules.
Two parallel encoders independently process PET and CT inputs to extract modality-specific features while sharing weights across branches to enhance computational efficiency. 
Each encoder consists of four Visual State Space (VSS) blocks with progressive down-sampling, enabling hierarchical feature abstraction at multiple spatial resolutions.

At each encoding stage, a Context-Gated Cross-Modal Perception Module (CGM) is introduced between the two branches to refine feature representations through adaptive gating. 
This mechanism dynamically highlights anatomically and functionally relevant regions by leveraging complementary contextual cues from both modalities, thereby improving inter-modality consistency and suppressing modality-specific noise.
Subsequently, the Dynamic Cross-Modality Interaction Module (DCIM)~\cite{mei2025cross} aggregates complementary information from both branches, promoting consistent feature alignment and effective fusion across modalities.

During decoding, the fused features are upsampled through Channel-Aware Visual State Space (CVSS) blocks~\cite{wan2025sigma}, which recover spatial details while preserving cross-modal consistency. 
The final output is then passed to a classifier to produce the tumor segmentation map.

\subsection{Context-Gated Cross-Modal Perception Module} 
Let $X_{\mathrm{CT}}, X_{\mathrm{PET}} \in \mathbb{R}^{C\times H\times W}$ denote feature maps extracted from CT and PET modalities, respectively. 
The Context-Gated Cross-Modal Perception Module (CGM) aims to enhance cross-modal representation learning by generating adaptive modulation masks conditioned on both modalities. 
This mechanism jointly captures global channel dependencies and local spatial correlations, allowing the model to emphasize informative regions and suppress modality-specific noise.  

Given the two inputs, the module first concatenates them along the channel dimension to form a fused tensor:
\begin{equation}
\mathbf{F} = [X_{\mathrm{CT}} ; X_{\mathrm{PET}}] \in \mathbb{R}^{2C\times H\times W}.
\end{equation}

A global average pooling operation aggregates contextual information from both modalities, producing:
\begin{equation}
\mathbf{p} = \operatorname{GAP}(\mathbf{F}) \in \mathbb{R}^{2C\times 1\times 1}.
\end{equation}

This pooled descriptor serves as input to a lightweight bottleneck composed of two $1\times1$ convolutions with batch normalization and GELU activation. 
For each modality \(m\in\{\mathrm{CT},\mathrm{PET}\}\), the channel gate is computed as:
\begin{equation}
\mathbf{c}_{m} = \sigma\!\left( \operatorname{Conv}^{(m)}_{1\times1}\left(\,\phi\!\left( \operatorname{BN}\!\left(\operatorname{Conv}^{(m)}_{1\times1}\,\left(\mathbf{p}\right)\right)\right)\right)\right),
\end{equation}
where $\phi$ and $\sigma$ denote the GELU and sigmoid functions, respectively, and $\mathbf{c}_{m} \in \mathbb{R}^{C\times1\times1}$ encodes the global channel importance of modality $m$.

To capture spatial correlations, the fused tensor $\mathbf{F}$ is also processed by a $3\times3$ convolution followed by a sigmoid activation, generating a spatial mask:
\begin{equation}
\mathbf{s}_{m} = \sigma\!\left( \operatorname{Conv}^{(m)}_{3\times3}(\mathbf{F}) \right),
\end{equation}
with $\mathbf{s}_{m} \in \mathbb{R}^{1\times H\times W}$.

The channel and spatial masks are combined multiplicatively to obtain the final gating tensor:
\begin{equation}
\mathbf{G}_{m} = \mathbf{c}_{m} \odot \mathbf{s}_{m},
\end{equation}
which modulates the input features of each modality through a residual scaling mechanism:
\begin{equation}
Y_{\mathrm{m}} = X_{\mathrm{m}} \odot (1 + \mathbf{G}_{\mathrm{m}}).
\end{equation}
Here, $\odot$ denotes element-wise multiplication with broadcasting along singleton dimensions.  

By learning asymmetric gates $\mathbf{G}_{\mathrm{CT}}$ and $\mathbf{G}_{\mathrm{PET}}$, each modality is enhanced according to complementary cues from the other, enabling CT-aware modulation of PET features and vice versa. 
This dual conditioning improves inter-modality coherence and reduces redundancy, while the residual formulation stabilizes training and preserves identity mappings when gates are near zero. 
Consequently, CGM adaptively emphasizes correlated anatomical--functional patterns and refines multimodal features to support accurate and context-aware tumor segmentation.

\section{Experimental setup}
Experiments were conducted on a workstation equipped with eight NVIDIA A40 GPUs. 
The model was trained for 50 epochs with a batch size of 8 using the AdamW optimizer with an initial learning rate of $6 \times 10^{-5}$, decayed through a cosine annealing schedule. 
Data augmentation techniques, including random horizontal and vertical flips and random cropping, were applied to enhance model generalization.

We evaluated the proposed method on the PCLT20K dataset, which contains 21,930 paired PET-CT slices from 605 lung cancer patients. Each pair is accompanied by a pixel-level tumor annotation generated through a three-stage procedure by experienced radiologists to ensure high-quality segmentation masks. 
CT images were clipped to \(\left[–1200, –200\right]\) Hounsfield Units to reduce background noise, while PET scans were converted to Standardized Uptake Value (SUV) maps before normalization. All images were resized to \(512 \times 512\) pixels for uniformity.

Performance was quantitatively assessed using three standard metrics: Intersection over Union (IoU), Dice coefficient, and the 95th percentile Hausdorff Distance (HD95).

\section{Results and Discussion}

\setlength{\fboxsep}{1pt}

\begin{table}[]
    \centering
    \resizebox{\columnwidth}{!}{
    \begin{tabular}{lccc|cc}
    \toprule
    \bfseries\makecell{Model} &
    \bfseries\makecell{IoU \(\bm{\uparrow}\) \\ (\%)} &
    \bfseries\makecell{Dice \(\bm{\uparrow}\) \\ (\%)} &
    \bfseries\makecell{HD95 \(\bm{\downarrow}\) \\ (mm) } &
    \bfseries\makecell{Flops \\ (G)} &
    \bfseries\makecell{Params \\ (M)} \\
    \midrule 
    SegResNet~\cite{myronenko20183d} & \(57.27\) & \(58.73\) &  \(37.25\) & 761.0 & 77.1 \\
    UNet~\cite{kerfoot2018left} & \(58.76\) & \(56.84\) &  \(57.33\) & 151.3 & 86.7 \\
    SwinUNETR~\cite{hatamizadeh2021swin} & \(55.05\) & \(57.44\) & \(22.85\) & 159.4 & 56.5 \\
    MedNeXt~\cite{roy2023mednext} & 58.95 & 58.77 & 25.49 & 277.5 & 41.0\\
    CIPA~\cite{mei2025cross} & \(59.56\) &  \(61.75\) & \(19.24\)& 80.2 & 54.6 \\
    \rowcolor{lightblue}
    vMambaX (Ours) & \(\bm{61.01}\) & \(\bm{61.96}\) & \(\bm{18.66}\) & 79.6 & 53.4 \\
    \bottomrule
    \end{tabular}
    }
    \caption{Comparison of segmentation performance and computational efficiency between our model and baselines. 
    The proposed model is highlighted in \colorbox{lightblue}{blue}.
    \textbf{Bold} values denote the best results.}
    \label{tab:quant_res}
\end{table}

\autoref{tab:quant_res} summarizes the quantitative comparison between vMambaX and state-of-the-art baselines.
Across all metrics, our model consistently outperforms competing methods while maintaining low computational complexity.
The improvements in both region- and distance-based metrics indicate more accurate lesion delineation and smoother, anatomically coherent boundaries.
These gains arise from the synergy between the Visual Mamba backbone and the CGM mechanism, which jointly enhance cross-modal feature integration and suppress modality-specific noise.

In terms of efficiency, vMambaX achieves higher segmentation performance with fewer parameters and FLOPs, offering a favorable balance between accuracy and computational cost.
Even compared to models with similar backbones, such as CIPA, vMambaX attains superior performance-efficiency trade-offs, showing that CGM improves feature interaction without additional overhead.

\begin{figure*}[!t]
\includegraphics[width=\textwidth]{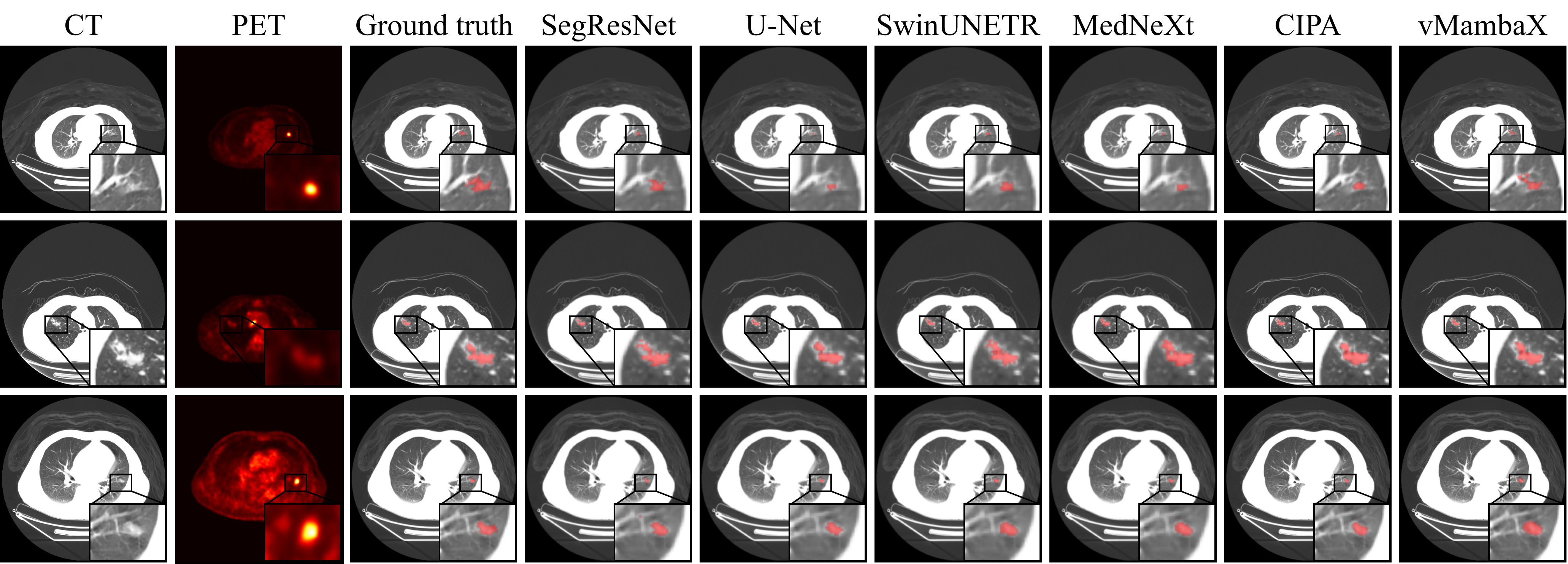}
\centering
\caption{Qualitative comparison of model segmentations, showing CT and PET inputs, ground truth, and predictions.}
\label{fig:qualitative_res}
\end{figure*}

\autoref{fig:qualitative_res} shows representative qualitative results on three PET-CT cases, demonstrating that vMambaX produces more precise and visually coherent segmentations, better capturing tumor extent and preserving boundary integrity.
While minor discrepancies persist along some contours, the model demonstrates improved robustness and fidelity to ground truth, confirming its qualitative advantage in integrating complementary CT and PET information.

\section{Conclusion}
\label{sec:conclusion}

This work presented vMambaX, a lightweight multimodal segmentation framework that integrates PET and CT information through a context-gated cross-modal perception mechanism.
Experiments on the PCLT20K dataset demonstrated that vMambaX achieves superior performance while maintaining low computational complexity, confirming the effectiveness of adaptive gating in leveraging complementary anatomical and functional cues.
Future work will extend the framework to 3D segmentation and assess its robustness across diverse datasets to further evaluate generalization.


\section{Compliance with ethical standards}
This study used publicly available anonymized imaging data; no ethical approval was required.

\section{Acknowledgments}
This work was partially supported by:
i) PNRR-MCNT2-2023-12377755, ii) CUP B89J23000580005, iii) AMP 23-1122, iv) CUP C83C25000210001 and v) JCSMK24-0094.
Computational resources were provided by NAISS and SNIC at Alvis @ C3SE, partially funded by the Swedish Research Council (grant nos. 2022-06725, 2018-05973).
The authors declare no competing interests.

\bibliographystyle{IEEEbib}
\bibliography{references}

\end{document}